\documentclass[runningheads]{llncs}
\usepackage{lipsum}
\usepackage{graphicx}
\usepackage{hyperref}
\makeatletter
\newcommand{\printfnsymbol}[1]{%
  \textsuperscript{\@fnsymbol{#1}}%
}
\makeatother

\begin{document}
\title{BERT-based Ensembles for Modeling Disclosure and Support in Conversational Social Media Text}
\titlerunning{BERT-based Ensembles for Modeling Disclosure and Support}
%
\author{Tanvi Dadu\inst{1} \printfnsymbol{1} \and 
Kartikey Pant\inst{2} \thanks{The first two authors contributed equally to the work.} \and 
Radhika Mamidi\inst{2}}
%
\authorrunning{Dadu, Pant and Mamidi}
%
\institute{Netaji Subhas Institute of Technology, Delhi \\
\email{tanvid.co.16@nsit.net.in}\\
\and 
International Institute of Information Technology, Hyderabad \\
\email{kartikey.pant@research.iiit.ac.in} \\
\email{radhika.mamidi@iiit.ac.in}
}
\maketitle              
\begin{abstract}
There is a growing interest in understanding how humans initiate and hold conversations. The affective understanding of conversations focuses on the problem of how speakers use emotions to react to a situation and to each other. In the CL-Aff Shared Task, the organizers released \textit{Get it \#OffMyChest} dataset, which contains Reddit comments from casual and confessional conversations, labeled for their disclosure and supportiveness characteristics. In this paper, we introduce a predictive ensemble model exploiting the finetuned contextualized word embeddings, \textit{RoBERTa} and \textit{ALBERT}. We show that our model outperforms the base models in all considered metrics, achieving an improvement of $3\%$ in the F1 score. We further conduct statistical analysis and outline deeper insights into the given dataset while providing a new characterization of impact for the dataset.

\end{abstract}
\section{Introduction}
The word \textit{‘Affective’} refers to emotions, mood, sentiment, personality, subjective evaluations, opinions, and attitude. Affect analysis refers to the techniques used to identify and measure the ‘experience of emotion’ in multimodal content containing text, audio, images, and videos.\cite{Rajendran2019HappyTL} Affect has become an essential part of the human experience, which directly influences their reaction towards a particular situation. Therefore, it has become crucial to analyze how speakers use emotions and sentiment to react to different situations and each other.

This paper addresses the challenge put forward in the CL-Aff Shared Task at the AAAI-2020 Workshop on Affective Content Analysis to Model Affect in Response (AffCon 2020). The theme of this task is to the study \emph{affect} in response to the interactive content which grows over time. The task offers two datasets ( a small labeled dataset and a large unlabeled dataset) sampled from casual and confessional conversations on Reddit in the subreddit \textit{/r/CasualConversations} and the \textit{/r/OffMyChest}. This shared task comprises two subtasks. The first subtask is a semi-supervised text classification task predicting \textit{Disclosure} and \textit{Supportiveness} labels based on the given two datasets. Whereas, the second subtask is an open-ended task, which requires authors to propose new characterizations and insights to capture conversation dynamics.

Recent works in the task of text classification have used pre-trained contextualized word representations rather than context-independent word representations. Some of these representations include BERT \cite{Devlin2019}, \textit{RoBERTa}\cite{2020roberta}, and \textit{ALBERT}\cite{2020albert}. These models perform contextualized word representation and are pre-trained using bidirectional transformers\cite{vaswaniattention}. These BERT-based pre-trained models have outperformed many existing techniques on most NLP tasks with minimal task-specific architectural changes. 

Ensemble models exploiting features learned from multiple pre-trained models are hypothesized to perform competitively. In this work, we propose an ensemble-based model exploiting pre-trained BERT-based word representations. We document the experimental results for the CL-Aff Shared Task of our proposed model in comparison to the baseline models. We further perform attribute-based statistical analysis using attributes like word count, day of the week, and comment per parent post. We conclude the paper by proposing impact as a new characterization to model conversation dynamics.
\section{Our Model}
In this section, we introduce our predictive model that uses Transfer learning in the form of pretrained BERT-based models. We propose an ensemble of two pre-trained models: \textit{RoBERTa} and \textit{ALBERT}. In this section, we first outline the pre-trained models incorporated and then discuss the ensemble technique used.

\subsection{Preliminaries}
Transfer learning is the process of extracting knowledge from a source problem domain and applying it to a different target problem or domain. Recent works on text classification use transfer learning in the form of pre-trained embeddings.\cite{yang2019xlnet,2020roberta,2020albert} These pre-trained embeddings have outperformed many of the existing techniques with minimal architectural structure. The use of pre-trained embeddings reduces the need for annotated data and allows one to perform the downstream task with minimal resources for the finetuning of the model.

Devlin et al.\cite{Devlin2019} introduced BERT, a contextualized word representation, pre-trained using a bi-directional Transformer-based encoder. These embeddings use a linear combination of masked language modeling and the next sentence prediction objectives. It is pre-trained on 3.3B words from various sources, including \textit{BooksCorpus}\cite{zhu2015aligning} and the \textit{English Wikipedia}. 

Liu et al. introduced \textit{RoBERTa}, a replication study of \textit{BERT}, with carefully tuned hyperparameters and more extensive training data\cite{2020roberta}. It is trained with a batch size eight times larger for half as many optimization steps, thus taking significantly lesser time to train in comparison. It is trained on more than twelve times the data used to train $BERT_{large}$, using data from \textit{OpenWebText} \cite{Gokaslan2019OpenWeb}, \textit{CC-News}\cite{CC-News}, and \textit{STORIES}\cite{journals/corr/abs-1806-02847} datasets. These optimizations lead the $RoBERTa_{large}$ pre-trained model to perform better than the BERT-large model in all benchmarking tests, including \textit{SQuAD}\cite{rajpurkar2016squad} and \textit{GLUE}\cite{wang-etal-2018-glue}. 

Lan et al. introduced \textit{ALBERT}, a BERT-based model with two parameter-reduction techniques: factorized embedding parameterization, and cross-layer parameter sharing.\cite{2020albert} These techniques help in lowering memory consumption and increasing training speed. Moreover, this model uses a self-supervised loss that focuses on modeling inter-sentence coherence and improves on downstream tasks with multi-sentence input. $ALBERT_{xxlarge,v2}$ achieves significant improvements over $BERT_{large}$ on multiple tasks.

\subsection{Our Approach}

\begin{figure}[!h]
        \center{\includegraphics[width=\textwidth]
        {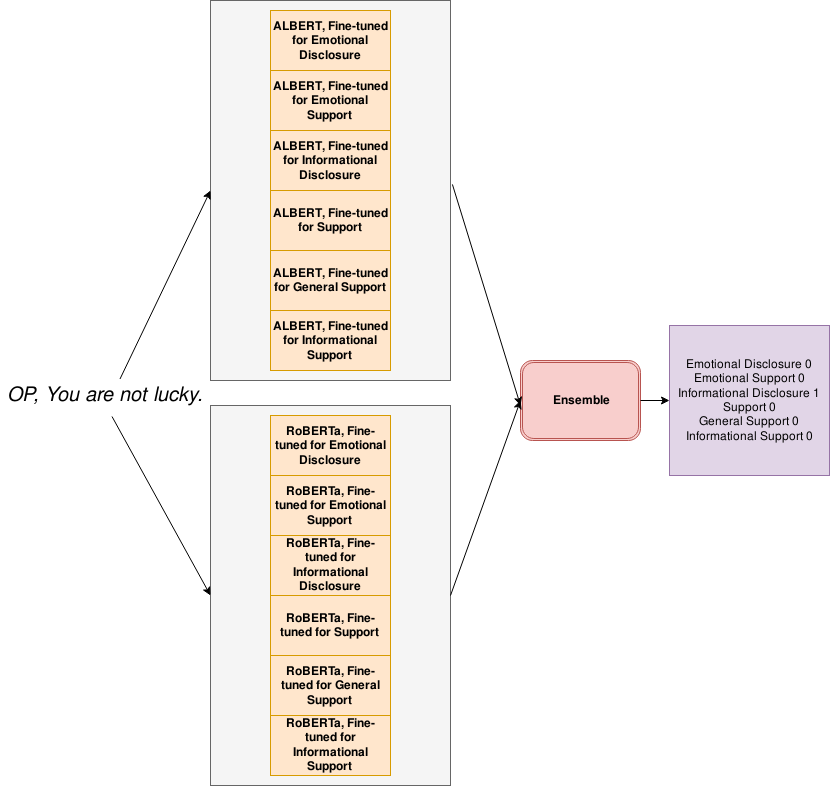}}
        \caption{\label{fig:1} Architecture of our ensemble models which predict the label set denoting support and disclosure from the comment text.}
\end{figure}

Ensemble methodology entails constructing a predictive model by integrating multiple models in order to improve prediction performance. They are meta-algorithms that combine several machine learning and deep learning classifiers into one predictive model to decrease variance, bias, and improve predictions. Recent works show that ensemble-based classifiers utilizing contextual embeddings outperform single-model classifiers.\cite{2020roberta,2020albert} Hence, we use ensembling techniques to combine predictions from multiple models for the tasks for making a prediction for the given task.

 \autoref{fig:1} depicts our proposed ensemble model. In this model, a sentence is parallelly computed by \textit{RoBERTa} and \textit{ALBERT} finetuned for predicting that label. The results from these base models are then combined using a weighted average based ensembling technique to predict the final label set, which includes predictions for the six labels.

\section{Experiments and Results}

In this section, we outline the experimental setup, the baselines for the task, and a comparative analysis of our proposed ensemble model with the two base models finetuned for the task, $RoBERTa_{large}$ and $ALBERT_{xxlarge,v2}$. We further compare our ensemble model with four other ensemble models and show that our model performs the best among all the models in four out of five evaluation metrics using 10-fold cross validation. 

For our baselines, we finetune $RoBERTa_{Large}$ and $ALBERT_{xxlarge,v2}$ models for three epochs with a maximum sequence length of $50$ and a batch size of $16$ for predicting each label separately. We finetune the model with a learning rate of $2*10^{-5}$, a weight decay of $0.01$, and $20$ steps for warm-up. We evaluate all the models on the following metrics: \textit{Accuracy}, \textit{F1}, \textit{Precision-1}, \textit{Recall-1}, and the mean of Accuracy and F1, denoted as \textit{Acc\&F1} from hereon.

\begin{table}[!h]
\centering
\resizebox{\textwidth}{!}{%
\begin{tabular}{|l|r|r|r|r|r|r|}
\hline
 \textbf{Model/Metrics} & \textbf{Accuracy} &  \textbf{Precision-1} &  \textbf{Recall-1} & \textbf{F1} & \textbf{Acc\&F1}  \\  \hline
 $RoBERTa_{Large}$ &  84.86\% &  0.585  & 0.514 &  0.541 & 0.695  \\ \hline
 $ALBERT_{xxlarge,v2}$  & 84.90\% & 0.596 & 0.472  & 0.524 & 0.686\\ \hline
 \textbf{Our Model}  & \textbf{85.55\%}  & \textbf{0.623} & \textbf{0.515} & \textbf{0.558} &  \textbf{0.707} \\  \hline
\end{tabular}%
}
\caption{Label-averaged values for each metric for \textit{RoBERTa},\textit{ALBERT}, and our best performing ensemble model.}
\label{table:1}
\end{table}

From \autoref{table:1}, we can discern that our ensemble-based model achieves the best results when compared with base models: \textit{RoBERTa} and \textit{ALBERT}. We observe a significant increase in \textit{Accuracy}, \textit{Precision-1}, and \textit{F1} and a slight increase in \textit{Recall-1} and \textit{Acc\&F1} in our best-performing ensemble model as compared to the base models.

\begin{table}[h!]
\centering
\resizebox{\textwidth}{!}{%
\begin{tabular}{|l|r|r|r|r|r|} \hline
 \textbf{Label/Metrics} & \textbf{Accuracy} &  \textbf{Precision-1} &  \textbf{Recall-1} & \textbf{F1} & \textbf{Acc\&F1}  \\  \hline
 \textbf{Informational Disclosure}  &  74.12\%  & 0.710 &  0.551 & 0.620  & 0.681 \\ \hline
 \textbf{Emotional Disclosure}  & 74.20\%  & 0.636 & 0.510 & 0.566 &  0.654\\ \hline
 \textbf{Support}  & 84.38\%   & 0.685 & 0.724 & 0.704 &  0.774 \\ \hline
 \textbf{General Support}  &  95.42\%  & 0.483 & 0.241 & 0.322 &  0.638\\ \hline
 \textbf{Informational Support}  & 91.30\%  & 0.592 & 0.485 &  0.533 & 0.723\\ \hline
 \textbf{Emotional Support}  & 93.86\% & 0.632 & 0.577 & 0.603 &  0.771\\ \hline
\end{tabular}%
}
\caption{Label-wise values for each metric for our best performing ensemble model.}
\label{table:2}
\end{table}

\autoref{table:2} further shows the performance of our ensemble-based model on individual labels. Its performance on different labels is evaluated using the above metrics. 

\begin{table}[h!]
\centering
\resizebox{\textwidth}{!}{%
\begin{tabular}{|l|r|r|r|r|r|}
\hline
 \textbf{Labels/Model} & \textbf{Model 1} & \textbf{Model 2} & \textbf{Model 3} & \textbf{Model 4} & \textbf{Model 5}  \\ \hline
 \textbf{Informational Disclosure} & 0.0,1.0 & 0.5,0.5 & 0.0,1.0 & 0.0,1.0 & 0.1,0.9   \\ \hline
 \textbf{Emotional Disclosure} &  0.0,1.0 & 0.5,0.5 & 0.5,0.5 & 0.5,0.5 & 0.5,0.5    \\ \hline
 \textbf{Support} & 1.0,0.0 & 0.5,0.5 & 1.0,0.0 & 1.0,0.0  &  1.0,0.0   \\ \hline
 \textbf{General Support} & 0.0,1.0 & 0.5,0.5 & 0.5,0.5 & 0.6,0.4 & 0.6,0.4   \\ \hline
 \textbf{Informational Support} & 1.0,0.0 & 0.5,0.5 & 1.0,0.0 & 1.0,0.0 & 1.0,0.0    \\ \hline
 \textbf{Emotional Support} & 1.0,0.0 & 0.5,0.5 & 0.5,0.5 & 0.5,0.5 & 0.5,0.5 \\ \hline
\end{tabular}%
}
\caption{Weights assigned to each model in different Ensemble Models. Each cell contains a pair $(x,y)$ where $x$ denotes the weight assigned to \textit{RoBERTa} and $y$ denotes the weight assigned to \textit{ALBERT}.}
\label{table:3}
\end{table}

We further performed a comparative study on ensembling techniques by choosing different weights for \textit{RoBERTa} and \textit{ALBERT}, as given in \autoref{table:3}. It shows different combinations of weights assigned to each label for \textit{RoBERTa} and \textit{ALBERT} respectively. This gives rise to five different models, which are then compared using the above metrics.

\begin{table}[h!]
\centering
\resizebox{\textwidth}{!}{%
\begin{tabular}{|l|r|r|r|r|r|r|} \hline
 \textbf{Model/Metrics} & \textbf{Accuracy} &  \textbf{Precision-1} &  \textbf{Recall-1} & \textbf{F1} & \textbf{Acc\&F1}  \\  \hline
 \textbf{Model 1}  & 85.18\%  & 0.595 &  \textbf{0.516} & 0.547  & 0.699 \\ \hline
 \textbf{Model 2}  & 85.42\%  & 0.622 & 0.490 & 0.544  & 0.699\\ \hline
 \textbf{Model 3}  & 85.47\%   & 0.619 & 0.514  & 0.557 &  0.706 \\ \hline
 \textbf{Model 4}  & 85.48\%  & 0.622  & 0.480 & 0.557  &  0.706\\ \hline
 \textbf{Model 5}  &  \textbf{85.54\%}  & \textbf{0.623} & 0.515 & \textbf{0.558} &  \textbf{0.707}\\ \hline
\end{tabular}%
}
\caption{Label-averaged values for each metric for different ensemble models.}
\label{table:4}
\end{table}

\autoref{table:4} depicts the results of the comparative study conducted on the five different ensemble models. We discern that \textit{Model 5} performs the best for \textit{Accuracy}, \textit{Precision-1}, \textit{F1}, and \textit{Acc\&F1} metrics, and \textit{Model 1} performs the best for \textit{Recall-1} metric among all the compared models. Since Model 5 outperforms all other models in four out of five metrics, it is the best predictive model for the task and is referred to as Our Model in the paper.

For the shared task, our \textit{System Run 1} to \textit{System Run 5} are predictions generated by the \textit{Model 1} to \textit{Model 5} respectively. \textit{System Run 6} and \textit{System Run 7} are the predictions generated by finetuned $RoBERTa\_{large}$ and $ALBERT_{xxlarge,v2}$ respectively.

\section{Dataset}
In this section, we provide a comprehensive statistical analysis of the dataset \textit{Get it \#OffMyChest}, which comprises of comments and parent posts from the subreddit /r/CasualConversations, and /r/OffMyChest. We further propose new characterizations and outline semantic features for the given dataset. 

\subsection{Analysis}
Statistical analysis of the labels, \textit{Emotional Disclosure}, \textit{Informational Disclosure}, \textit{Support}, \textit{General Support}, \textit{Information Support}, and \textit{Emotional Support} show significant variations in the number of positive and negative labels. The percentage of positive labels is maximum for \textit{Information Disclosure} with $37.99\%$ and minimum for \textit{General Support} with $5.37\%$. Therefore, the given dataset is highly imbalanced, which makes the training of predictive models a strenuous task.

Further analysis of the labeled dataset shows that there are $3,511$ parent posts for $11,573$ comments. We observe an average of $3.29$ comments per parent post ranging from one comment per parent post to $52$ comments per parent post. In the given dataset, there are $6,999$ unique users with an average of $1.653$ comments per user and a significant variation in the number of comments per user ranging from $1$ to $159$ comments per user, with a standard deviation of $2.669$. From this, we conclude that multiple comments within the same parent post and by the same author may be related to each other.
\begin{table}[]
\centering
\resizebox{\textwidth}{!}{%
\begin{tabular}{|l|r|r|r|r|r|r|}
\hline
\textbf{Weekday/Label} & \multicolumn{1}{l|}{\textbf{\begin{tabular}[c]{@{}l@{}}Emotional\\ Disclosure\end{tabular}}} & \multicolumn{1}{l|}{\textbf{\begin{tabular}[c]{@{}l@{}}Informational\\ Disclosure\end{tabular}}} & \multicolumn{1}{l|}{\textbf{Support}} & \multicolumn{1}{l|}{\textbf{\begin{tabular}[c]{@{}l@{}}General\\ Support\end{tabular}}} & \multicolumn{1}{l|}{\textbf{\begin{tabular}[c]{@{}l@{}}Informational\\ Support\end{tabular}}} & \multicolumn{1}{l|}{\textbf{EmotionalSupport}} \\ \hline
Monday                 & 29.73\%                                                                                      & 38.55\%                                                                                          & 25.70\%                               & 6.01\%                                                                                  & 9.49\%                                                                                        & 7.89\%                                         \\ \hline
Tuesday                & 29.71\%                                                                                      & 38.23\%                                                                                          & 24.80\%                               & 5.43\%                                                                                  & 9.66\%                                                                                        & 7.26\%                                         \\ \hline
Wednesday              & 30.70\%                                                                                      & 38.35\%                                                                                          & 26.80\%                               & 5.93\%                                                                                  & 11.68\%                                                                                       & 7.79\%                                         \\ \hline
Thursday               & 29.40\%                                                                                      & 37.27\%                                                                                          & 24.95\%                               & 5.08\%                                                                                  & 9.78\%                                                                                        & 7.75\%                                         \\ \hline
Friday                 & 31.14\%                                                                                      & 35.53\%                                                                                          & 24.67\%                               & 5.27\%                                                                                  & 8.41\%                                                                                        & 8.35\%                                         \\ \hline
Saturday               & 30.59\%                                                                                      & 38.95\%                                                                                          & 22.04\%                               & 4.34\%                                                                                  & 8.22\%                                                                                        & 6.32\%                                         \\ \hline
Sunday                 & 31.85\%                                                                                      & 38.92\%                                                                                          & 25.93\%                               & 5.41\%                                                                                  & 10.31\%                                                                                       & 9.06\%                                         \\ \hline
\textbf{Overall}       & \textbf{30.44\%}                                                                             & \textbf{37.99\%}                                                                                 & \textbf{25.02\%}                      & \textbf{5.37\%}                                                                         & \textbf{9.66\%}                                                                               & \textbf{7.79\%}                                \\ \hline
\end{tabular}%
}
\caption{Weekday-wise label distribution of the labelled dataset.}
\label{table:5}
\end{table}

We also observe significant variations in the word count of the comments, with an average comment being of $14.7$ words, which translates to around one sentence\cite{SentenceLengthPaper}. However, the comment length varies significantly from $3$ words to $151$ words per comment, with the distribution having a standard deviation of $9.670$. The dataset is thus, well-rounded, and represents realistic discourse setting with participants exchanging comments of varying lengths.

We intuitively proceeded to predict the effect of the day of the week in the characterized labels representing disclosure and support in a comment. It was expected that the users would behave differently as the week progresses. However,  as illustrated in \autoref{table:5}, we do not see any significant variation in the existing characterizations with a change in the day of the week. Thus, we conclude, in this dataset, that the week of the day doesn’t affect the users to be either more supportive or disclose more information.

\subsection{Impact Prediction}

The score assigned to a comment quantifies its \textit{Impact} since, on Reddit, it is the difference between the upvotes and downvotes that it obtains. We observed the posts to have a moderately positive \textit{Impact} of $10.938$ on average. We also see that the breadth of the spectrum in the \textit{Impact} is captured well by the dataset, with a standard deviation of $57.198$, and a range of $-49$ to $2,637$. This paves the way for a need to characterize and predict the \textit{Impact} of a post.

\begin{table}[]
\centering
\resizebox{0.5\textwidth}{!}{%
\begin{tabular}{|l|r|}
\hline
\textbf{Labels} & \textbf{$\rho$ with Impact}  \\ \hline
\textbf{Emotional Disclosure} & 0.046 \\ \hline
\textbf{Informational Disclosure} & 0.024 \\ \hline
\textbf{Support} &  0.021 \\ \hline
\textbf{General Support} & 0.028 \\ \hline
\textbf{Information Support} & 0.005 \\ \hline
 \textbf{Emotional Support}&  0.019 \\ \hline
\end{tabular}%
}
\caption{The relationship between Labels and Impact, as represented by Pearson correlation coefficient, $\rho$.}
\label{table:6}
\end{table}
Upon performing a correlation study between Impact and the previously characterized labels using \textit{Pearson’s correlation coefficient}\cite{Rodgers1988}, we observe a very small positive correlation between the variables. As is illustrated in \autoref{table:6}, the maximum $\rho$ of $0.046$ between \textit{Impact} and \textit{Emotional Disclosure} represents that \textit{Impact} is characteristically distinct from the previously predicted labels. 

We further analyze the influence of \textit{Impact}, characterized by the score on the semantic structure of the comments. We perform a correlation study between \textit{Impact} and semantic features selected, as is explored previously in Yang et al\cite{HumorRecognitionPaper}. Semantic structure is captured by the following features:  
\begin{enumerate}
    \item \textbf{Positive words}: The number of occurrences of positive words in a comment.
    \item \textbf{Negative words}: The number of occurrences of negative words in a comment.
    \item \textbf{Positive Polarity Confidence}: The probability that a sentence is positive. This metric is used to capture the polarity of comments and is calculated using Fasttext\cite{bojanowski2016enriching}.
    \item \textbf{Subjective words}: The number of occurrences of subjectivity oriented words in a comment. It is used to capture the linguistic expression of people’s opinions, beliefs, and speculations.
    \item \textbf{Sense Combination}: It is computed as the  $log( \Pi^{k}_{i=1} n_{wi} )$ where $n_{wi}$ is the total number of senses of word $w_{i}$.
    \item \textbf{Sense Farmost}: The largest Path Similarity of any word sense in a sentence.
    \item \textbf{Sense Closest}: The smallest Path Similarity of any word sense in a sentence.
\end{enumerate}

\begin{table}[]
\centering
\resizebox{0.5\textwidth}{!}{%
\begin{tabular}{|l|r|}
\hline
\textbf{Semantic Features} & \textbf{$\rho$ with Impact}  \\ \hline
\textbf{Positive words} & -0.009 \\ \hline
\textbf{Negative words} & -0.018 \\ \hline
\textbf{Subjective words} &  -0.018 \\ \hline
\textbf{Sense Combination} & -0.019 \\ \hline
\textbf{Sense Farmost} & -0.015 \\ \hline
\textbf{Sense Closest}&  0.040 \\ \hline
\textbf{Positive Polarity Confidence}&  -0.012 \\ \hline
\end{tabular}%
}
\caption{The relationship between Semantic Features and Impact, as represented by Pearson correlation coefficient, $\rho$.}
\label{table:7}
\end{table}

From \autoref{table:7}, we observe a minimal correlation between \textit{Impact} and the selected semantic features. The maximum $\rho$ of $0.040$ between \textit{Impact} and the feature \textit{Sense Closest}\cite{HumorRecognitionPaper} depicts that the new characterization is distinct from semantic features of the comment. 

Although it is essential to understand that predicting \textit{Impact} is beneficial for numerous applications like finance, product marketing and provides insights on social dynamics, it is a hard problem dependent on various factors. Our attempt to capture relationships between \textit{Impact} and some selected semantic features was not able to establish a strong correlation between the features. Thus, this implies that the use of sophisticated architectures in the task of \textit{Impact} prediction would be valuable. 

\section{Conclusion}
This paper presents a novel BERT-based predictive ensemble model to predict given labels: \textit{Emotional Disclosure}, \textit{Informational Disclosure}, \textit{Support}, \textit{General Support}, \textit{Information Support}, and \textit{Emotional Support}. Our model gives competitive results for the label prediction on the given dataset \textit{Get it \#OffMyChest}. Analysis of dataset shows the highly imbalanced distribution of the given labels, and high variations in some features like score, word count, comments per parent post, and comments per user. We further discerned that day of the week has no significant impact on the frequency of Disclosure and Support based comments on Reddit. Future work may involve exploring more ensembling techniques and exploring sophisticated architectures to predict the impact of a comment.

%
%
%
\bibliographystyle{splncs04}
\bibliography{affcon-paper}

\begin{thebibliography}{10}
\providecommand{\url}[1]{\texttt{#1}}
\providecommand{\urlprefix}{URL }
\providecommand{\doi}[1]{https://doi.org/#1}

\bibitem{bojanowski2016enriching}
Bojanowski, P., Grave, E., Joulin, A., Mikolov, T.: Enriching word vectors with
  subword information (2016), \url{http://arxiv.org/abs/1607.04606}

\bibitem{SentenceLengthPaper}
Cutts, M.: Oxford Guide to Plain English (2009)

\bibitem{Devlin2019}
Devlin, J., Chang, M.W., Lee, K., Toutanova, K.: Bert: Pre-training of deep
  bidirectional transformers for language understanding (2018),
  \url{http://arxiv.org/abs/1810.04805}

\bibitem{Gokaslan2019OpenWeb}
Gokaslan, A., Cohen, V.: Openwebtext corpus.
  \url{http://Skylion007.github.io/OpenWebTextCorpus} (2019)

\bibitem{CC-News}
Nagel, S.: Cc-news (2016),
  \url{http://web.archive.org/save/http://commoncrawl.org/2016/10/newsdataset-available/}

\bibitem{Rajendran2019HappyTL}
Rajendran, A., Zhang, C., Abdul-Mageed, M.: Happy together: Learning and
  understanding appraisal from natural language. In: AffCon@AAAI (2019)

\bibitem{rajpurkar2016squad}
Rajpurkar, P., Zhang, J., Lopyrev, K., Liang, P.: Squad: 100,000+ questions for
  machine comprehension of text. arXiv preprint arXiv:1606.05250  (2016)

\bibitem{Rodgers1988}
Rodgers, J., Nicewander, W.: Thirteen ways to look at the correlation
  coefficient. The American Statistician  \textbf{42}(1),  59--66 (1988),
  \url{http://www.jstor.org/stable/2685263}

\bibitem{journals/corr/abs-1806-02847}
Trinh, T.H., Le, Q.V.: A simple method for commonsense reasoning. CoRR
  \textbf{abs/1806.02847} (2018),
  \url{http://dblp.uni-trier.de/db/journals/corr/corr1806.html#abs-1806-02847}

\bibitem{vaswaniattention}
Vaswani, A., Shazeer, N., Parmar, N., Uszkoreit, J., Jones, L., Gomez, A.N.,
  Kaiser, {\L}., Polosukhin, I.: Attention is all you need. In: Advances in
  Neural Information Processing Systems. pp. 5998--6008 (2017)

\bibitem{wang-etal-2018-glue}
Wang, A., Singh, A., Michael, J., Hill, F., Levy, O., Bowman, S.: {GLUE}: A
  multi-task benchmark and analysis platform for natural language
  understanding. In: Proceedings of the 2018 {EMNLP} Workshop {B}lackbox{NLP}:
  Analyzing and Interpreting Neural Networks for {NLP}. pp. 353--355.
  Association for Computational Linguistics, Brussels, Belgium (Nov 2018).
  \doi{10.18653/v1/W18-5446}, \url{https://www.aclweb.org/anthology/W18-5446}

\bibitem{HumorRecognitionPaper}
Yang, D., Lavie, A., Dyer, C., Hovy, E.H.: Humor recognition and humor anchor
  extraction. In: Màrquez, L., Callison-Burch, C., Su, J., Pighin, D., Marton,
  Y. (eds.) EMNLP. pp. 2367--2376. The Association for Computational
  Linguistics (2015),
  \url{http://dblp.uni-trier.de/db/conf/emnlp/emnlp2015.html#YangLDH15}

\bibitem{yang2019xlnet}
Yang, Z., Dai, Z., Yang, Y., Carbonell, J., Salakhutdinov, R., Le, Q.V.: Xlnet:
  Generalized autoregressive pretraining for language understanding (2019),
  \url{http://arxiv.org/abs/1906.08237}

\bibitem{2020roberta}
Yinhan~Liu, Myle~Ott, N.G.J.D.M.J.D.C.O.L.M.L.L.Z.V.S.: Roberta: A robustly
  optimized bert pretraining approach. In: Submitted to International
  Conference on Learning Representations (2020),
  \url{https://openreview.net/forum?id=SyxS0T4tvS}, under review

\bibitem{2020albert}
Zhenzhong~Lan, Mingda~Chen, S.G.K.G.P.S.R.S.: Albert: A lite bert for
  self-supervised learning of language representations. In: Submitted to
  International Conference on Learning Representations (2020),
  \url{https://openreview.net/forum?id=H1eA7AEtvS}, under review

\bibitem{zhu2015aligning}
Zhu, Y., Kiros, R., Zemel, R., Salakhutdinov, R., Urtasun, R., Torralba, A.,
  Fidler, S.: Aligning books and movies: Towards story-like visual explanations
  by watching movies and reading books (2015),
  \url{http://arxiv.org/abs/1506.06724}

\end{thebibliography}

\end{document}